\documentclass[10pt,twocolumn,letterpaper]{article}
\usepackage{cvpr}              

\usepackage{graphicx}
\usepackage{amsmath}
\usepackage{amssymb}
\usepackage{booktabs}
\usepackage{algorithm}
\usepackage{algorithmic}
\usepackage{multirow}
\usepackage{color}
\usepackage{ulem}
\usepackage{tabularx}
\usepackage{subcaption,float}
\usepackage{amssymb}
\usepackage{pifont}
\newcommand{\cmark}{\ding{51}}%
\newcommand{\xmark}{\ding{55}}%

\usepackage[pagebackref,breaklinks,colorlinks]{hyperref}

\usepackage[capitalize]{cleveref}
\crefname{section}{Sec.}{Secs.}
\Crefname{section}{Section}{Sections}
\Crefname{table}{Table}{Tables}
\crefname{table}{Tab.}{Tabs.}

\def\cvprPaperID{7259} 
\def\confName{CVPR}
\def\confYear{2022}

\newif\ifcomments
\commentstrue 

\definecolor{blue}{rgb}{0,0.1,0.9}

\definecolor{red}{rgb}{0.9,0.1,0}

\begin{document}

\title{Dressing in the Wild by Watching Dance Videos\vspace{-2mm}}

\author{
Xin Dong{$^{1}$},
~ Fuwei Zhao{$^{2}$},
~ Zhenyu Xie{$^{2}$},
~ Xijin Zhang{$^{1}$}\\\vspace{-16pt}\\ 
Daniel K. Du{$^{1}$},
~ Min Zheng{$^{1}$},
~ Xiang Long{$^{1}$},
~ Xiaodan Liang{$^{2*}$},
~ Jianchao Yang{$^{1}$}\\\vspace{-10pt}\\
{$^{1}$}ByteDance, {$^{2}$}Shenzhen Campus of Sun Yat-Sen University\\
\small{\tt{\{zhaofw@mail2,xiezhy6@mail2,xdliang328@mail\}.sysu.edu.cn}}\\ \small{\tt{\{dongxin.1016,zhangxijin,dukang.daniel,zhengmin.666,longxiang.0,yangjianchao\}@bytedance.com}}
\vspace{-2mm}
}

\vspace{-10mm}
\twocolumn[{%
\renewcommand\twocolumn[1][]{#1}%
\maketitle
    \includegraphics[width=1.0\hsize]{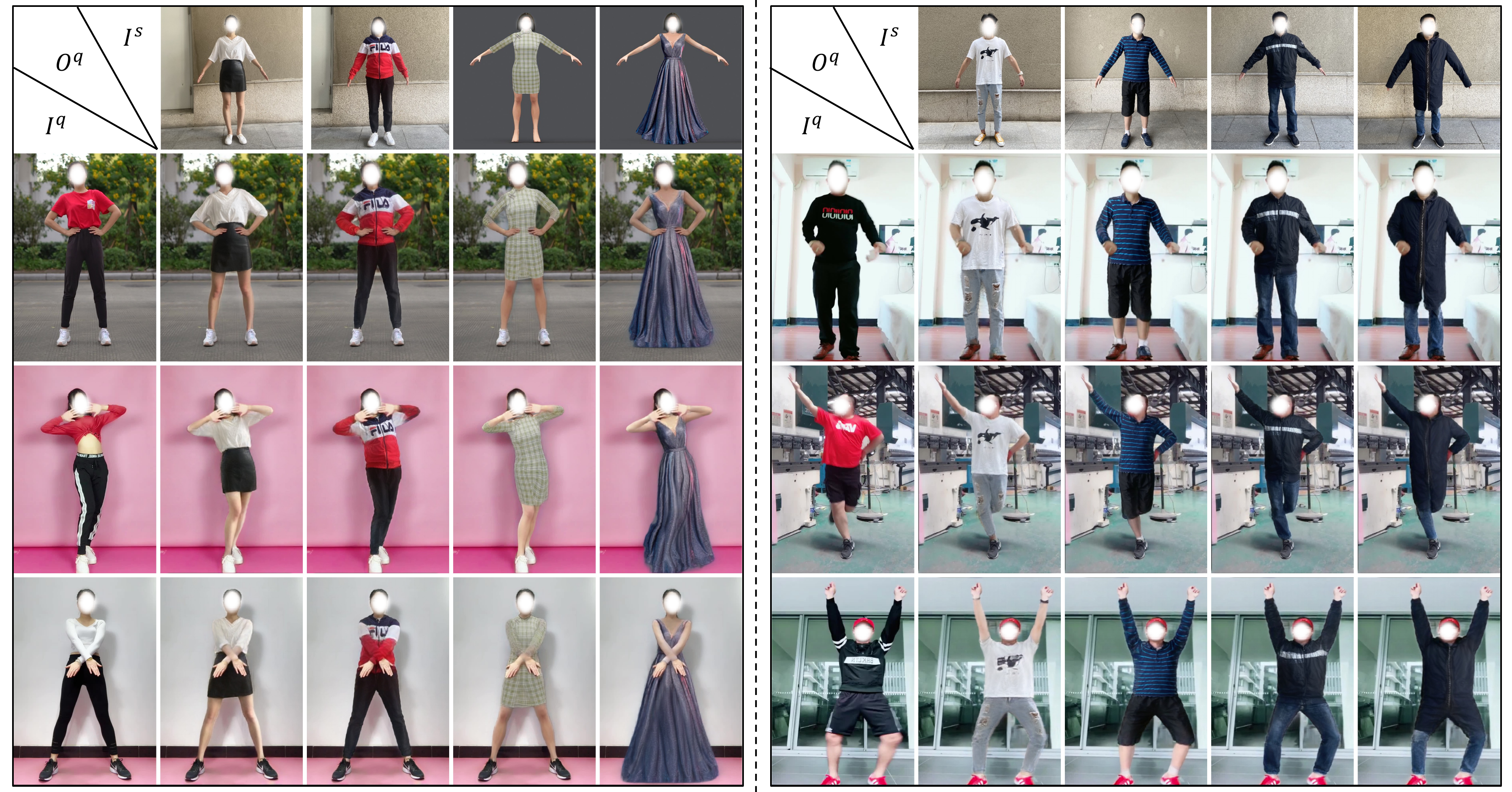}
    \vspace{-6mm}
    \captionof{figure}{The results of the proposed method on in-the-wild images. Our model is capable of transferring arbitrary garments (e.g., shirts, pants, formal dresses, skirts, down jackets) from a source person image $I^s$ onto a challenging posed query person image $I^q$ presented in real-world backgrounds, generating high-fidelity output image $O^q$ with the query's identity now wearing the source's garments.\newline\newline} 
    \vspace{-4mm}
    \label{fig:teaser}
}]

\begin{abstract}
\vspace{-6mm}
While significant progress has been made in garment transfer, one of the most applicable directions of human-centric image generation, existing works overlook the in-the-wild imagery, presenting severe garment-person misalignment as well as noticeable degradation in fine texture details. This paper, therefore, attends to virtual try-on in real-world scenes and brings essential improvements in authenticity and naturalness especially for loose garment (e.g., skirts, formal dresses), challenging poses (e.g., cross arms, bent legs), and cluttered backgrounds. Specifically, we find that the pixel flow excels at handling loose garments whereas the vertex flow is preferred for hard poses, and by combining their advantages we propose a novel generative network called wFlow that can effectively push up garment transfer to in-the-wild context. Moreover, former approaches require paired images for training. Instead, we cut down the laboriousness by working on a newly constructed large-scale video dataset named \textit{Dance50k} with self-supervised cross-frame training and an online cycle optimization. The proposed \textit{Dance50k} can boost real-world virtual dressing by covering a wide variety of garments under dancing poses. Extensive experiments demonstrate the superiority of our wFlow in generating realistic garment transfer results for in-the-wild images without resorting to expensive paired datasets.~\footnote{Xiaodan Liang is the corresponding author. The project page of wFlow is ~\url{https://awesome-wflow.github.io}.}
\end{abstract}

\vspace{-8mm}
\section{Introduction}
\label{sec:intro}

Garment transfer, the process of transferring garments onto a query person image without changing the person's identity, is a central problem in human-centric image generation that promises great commercial potential. However, when getting down to in-the-wild scenarios, general solutions are required that can leverage easily accessible training data, handle arbitrary garments, and cope with complex poses presented in real world environments. 

Unluckily, most existing works~\cite{han2018viton,Matiur2020cpvton+,wang2018cpvton,yu2019vtnfp,m3dvton,yang2020acgpn,Ge_2021_dcton,pf-afn,Xie2021WASVTONWA} serve to fit an in-shop garment to a target person by utilizing either pixel flow~\cite{han2019clothflow} or TPS transformation~\cite{bookstein1989TPS}. Despite their promise, these methods become less effective at exchanging garments directly between two persons, due to the deficiency of the 2D transformations when faced with large pose variations. Also, previous methods require paired data for training, i.e., a person and its associated garment image, which further leads to laborus collection process. These limitations largely hinder their practical use and raise the need for scalable solutions that can be trained on easily accessible data. 
\cite{Liu_2019_liquid} takes a step forward by replacing the 2D pixel flow with 3D SMPL~\cite{matthew2015smpl} vertex flow, allowing person-to-person garment transfer and can address complex poses or severe self-occlusion. However, it is error prone to loose garments that can not be modeled as part of the SMPL surface. Albeit subjects to simple poses, the 2D pixel flow can then again predict more faithful pixel mapping for these challenging loose garments.

Therefore, in this paper, we propose \textit{wFlow} that efficiently integrates respective advantages of the 2D pixel flow and the 3D vertex flow. Based on the wFlow, a robust garment transfer network is developed to tackle the essential challenges on in-the-wild imagery. In particular, we design a self-supervised training scheme that works on easily obtainable dance videos by exploiting cross-frame consistency, getting rid of the hard-to-get paired dataset.

Our insight is that a well-designed flow-based model trained on multi-pose images of the same person, which is easily accessible in dance videos, can generalize well at testing to transfer garments between different persons by adding protected body parts that guides the network to focus on the garment regions. Thus, we collect thousands of single-person dance videos with diverse garment types as the training dataset, and sample from it a plethora of multi-pose person frames to train our model in a similar fashion of pose transfer~\cite{Li_2019_dense,Neverova2018DensePT}, where the designed wFlow that associates different frames allows us to self-supervise the training procedure without ground truth flow supervision.

To fully exploit the wFlow, we first pass the source and query person representation through a conditional segmentation network, producing a person segmentation that complies with both the source garment and the query pose. Given the predicted segmentation, a pixel flow network is employed to estimate the pixel-wise correspondences between the source and the query images. Thereafter, we compute 3D SMPL vertex flow directly from the inputs and project it to image plane where the pixel flow is also injected to form the proposed wFlow. The warped garment can now be obtained by applying the wFlow on the source garment. Finally, a skip-connected inpainting network leverages the warped garment along with the protected person regions and fuses them with the inpainted query background. Additionally, for garments presented scarcely in training data, we further formulate a cyclic online optimization to enhance the quality of their transfer results. Thanks to the contributory video data and the potent texture mapping ability of the wFlow, our model can handle arbitrary garments and seamlessly transfer them onto challenging posed query persons.

Overall, we present three main contributions:
\begin{itemize}
    \item We are the first to explore the in-the-wild garment transfer problem. By exploiting a self-supervised training scheme that works on easily accessible dance videos, our model generates surprising results with sharp textures and intact garment shape.
    \item To facilitate arbitrary garment transfer under complex poses in real-world scenario, we introduce a novel \textbf{wFlow} (flow in-the-wild) that integrates both 2D and 3D information along with a cyclic online optimization that further enhances the synthesis quality.
    \item We construct a new large-scale video dataset called \textit{Dance50k}, containing 50k sequences of dancing people wearing a wide variety of garments, which is useful for the development of human-centric image/video processing not limited to virtual try-on.
\end{itemize}

\section{Related Work}
\label{sec:related}
\subsection{Image-based Garment Transfer}
Due to the great application potential, research on 2D garment transfer has been explored intensively~\cite{la-vton, Matiur2020cpvton+,minar2020clothviton,choi2021vitonhd,Ge_2021_dcton,dong2019mgvton,Xie2021WASVTONWA,Wang2020DownTT,scvton,albahar2021pose}. 
VITON~\cite{han2018viton} and CP-VTON~\cite{wang2018cpvton} are the starting point in this convincing field. Both of them utilize a TPS-based deformation module followed by a texture fusion module to warp and fuse a catalog garment to a query person image.
VTNFP~\cite{yu2019vtnfp} and ACGPN~\cite{yang2020acgpn} inherit the same warping scheme but further introduce the human parsing as synthesis guidance, achieving better delineation at cloth-skin boundary. More recently, PF-AFN~\cite{pf-afn} gets rid of human parsing by exploiting a appearance flow distillation scheme, generating consistently good results via a simpler student model. 
Despite their promise, all these methods require a training set consisting of paired images. This limits the scale at which training data can be collected since obtaining such paired images is highly laborious. Also, during testing only in-shop garments can be transferred to the query persons with simple poses. 

To resolve this, \cite{neuberger2020oviton,lewis2021tryongan,Sarkar2020NeuralRO,sarkar2021styleposegan,Sarkar2021HumanGAN, men2020adgan, pastagan} extend the paired approaches to their unpaired counterparts that realize garment transfer directly between two persons. Most of these works rely on the powerful conditional generator (e.g., StyleGAN2~\cite{Karras_2020_stylegan2}) to manipulate the latent garment feature embeddings.
However, unlike the flow warping that directly transforms textures, these pure generation models inevitably suffer with recovering complex texture patterns from the low-dimensional latent space, even with the state-of-the-art StyleGAN2. Overall, previous garment transfer methods, are either limited by the short-supplied data or the insufficient GAN inversion. Our method, instead, can not only leverage in-the-wild video data but also yield realistic garment and skin textures. In the next subsection, we will give a brief review on the flow estimation that we adapt for modeling the texture mapping mechanism.

\subsection{Human-centric Flow Estimation}
Optical flow, defined for representing pixel offsets between adjacent frames in videos~\cite{Dosovitskiy2015FlowNet,Ilg2017FlowNet2,Teed2020RAFT,Zhao2020MaskFlownet,Sun2018PWCNetCF,Jonschkowski2020unsupervisedflow}, has also inspired researchers in the field of human-centric generation for tackling problems such as face hallucination~\cite{Song2019JointFace}, pose transfer~\cite{Li_2019_dense}, and virtual try-on~\cite{han2019clothflow,pf-afn,Dong2019FWGANFW}. Without loss of meaning, we rename the optical flow as \textbf{2D pixel flow} for human generation, which refers to 2D coordinate vectors indicating which pixels in the source can be used to synthesize the target. ClothFlow~\cite{han2019clothflow} and FW-GAN~\cite{Dong2019FWGANFW} are the firsts to utilize the pixel flow for garment transfer. Thanks to its high degrees of freedom, 2D pixel flow is suitable for dressing both skin-tight and loose garments, however, it tends to fail when faced with large pose variations due to the ignorance of underlying rigid body information. \cite{Liu_2019_liquid}, \cite{Yoon_2021_CVPR} and \cite{Li_2019_dense} thus turn to \textbf{3D vertex flow} based on 3D SMPL model~\cite{matthew2015smpl} that is kinematics-aware to cope with complex poses.
While outperform on diverse poses, these methods sacrifice the deformation freedom compared to the pixel flow, generating unsatisfactory results of loose clothes. Given clear advantages of both sides, we make the first attempt to leverage both 2D pixel flow and 3D vertex flow to facilitate in-the-wild virtual dressing.

\section{Methodology}
\begin{figure*}
\begin{center}
\includegraphics[width=1.0\hsize]{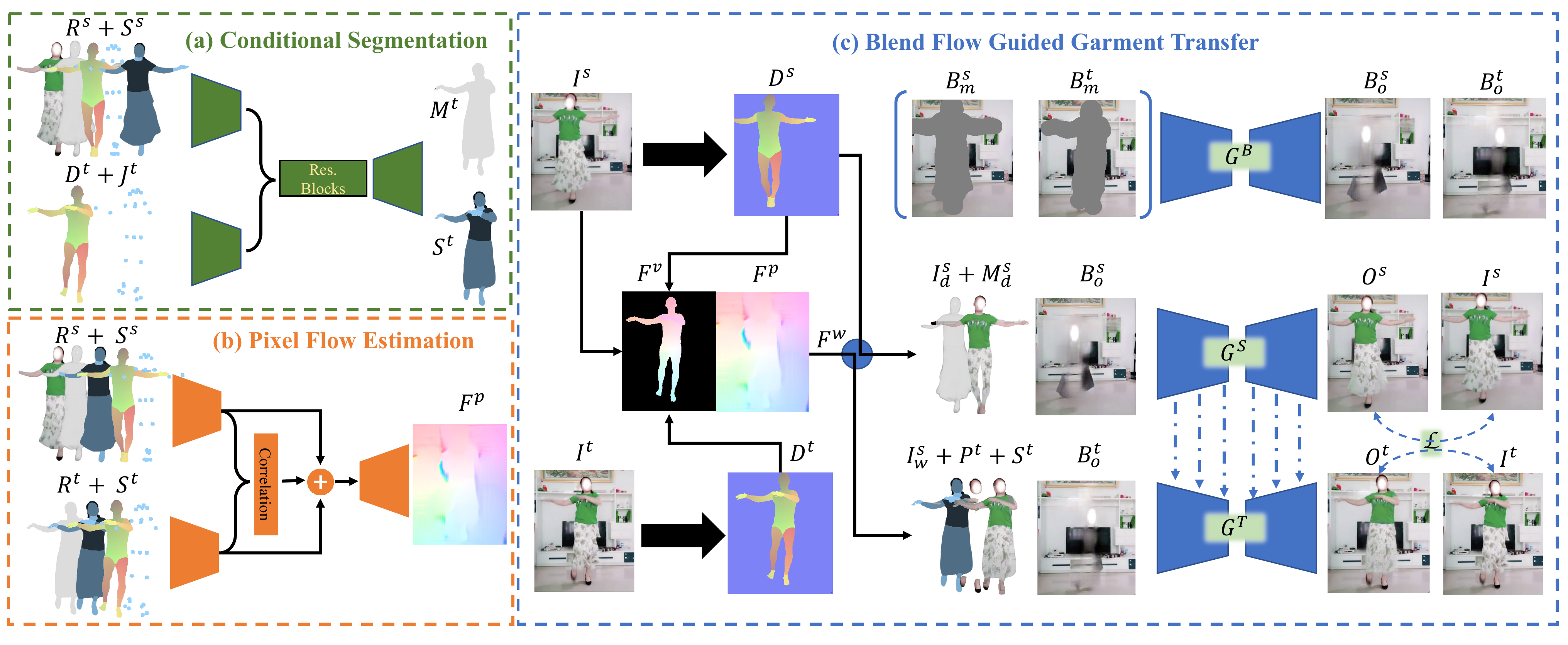}
\end{center}
    \vspace{-6mm}
   \caption{Architecture Overview. Our training pipeline contains: (a) Conditional Person Segmentation (Sec.~\ref{subsection:stage1}): generates the person layouts $\left(M^t,S^t\right)$ providing the source garment shape under the target pose; (b) Pixel Flow Estimation (Sec.~\ref{subsection:stage2}): leverages $\left(M^t,S^t\right)$ and other person representations to estimate a per-pixel flow map $F^p$. (c) wFlow-Guided Garment Transfer (Sec.~\ref{subsection:stage3}): computes first the 3D vertex flow $F^v$ from the source and target person frame $\left(I^s,I^t\right)$ and fuses it with $F^p$ to form the wFlow $F^w$. Thereafter the warped person $I^s_w$ is fitted to the $G^B$-inpainted background $B^t_o$ by a feature distillation generator $G^T$ that incorporates the texture completion ability of the reconstruction generator $G^S$, producing respectively the output $O^t$/$O^s$ supervised by $I^t$/$I^s$.}
  \vspace{-4mm}
\label{fig:framework}
\end{figure*}

As shown in Fig.~\ref{fig:teaser}, given a source image $I^s$ of a person and a query image $I^q$, the goal of our method is to generate a synthetic image $O^q$ preserving the query's identity and background but now dressing the source's garment. While training the model with $\left(I^s,I^q,O^q\right)$ triplets is straightforward, collecting such a dataset is beyond laborious since $O^q$ is usually unavailable. Instead, we use $\left(I^s,I^t,O^t\right)$ where the query (hereafter denoted as \textit{target} for the training phase) and the source are the same person under different poses, which is easily accessible from different frames in dance videos. Fig.~\ref{fig:framework} illustrates the overall architecture that exploits the designed wFlow in a multi-stage manner described in the following subsections.

\subsection{Stage 1: Conditional Person Segmentation}\label{subsection:stage1}
Our training approach in summary is similar to the pose transfer prototype~\cite{Li_2019_dense,Liu_2019_liquid}. However, directly extending pose transfer to garment transfer is prone to overfitting, seeing that during training we use multi-pose images of the same person while at testing the query can be arbitrary identity. To provide more reliable synthesis guidance for inference, a Conditional Segmentation Network (CSN) is employed to predict person segmentation layout that conforms to the target shape as well as the source garment, as shown in Fig.~\ref{fig:framework}.(a).

Specifically, with two separate encoders, the CSN first extracts features from two image collections respectively: (1) the 20-channel source person segmentation $S^s$ (obtained by applying~\cite{liang2018look} on the source image $I^s$) together with the person representation $R^s$ including a 3-channel RGB image, a 1-channel person mask (obtained by binarizing $S^s$), a 3-channel densepose (obtained by projecting fitted SMPL mesh~\cite{kanazawaHMR18} to 2D UV space), and a 18-channel body joints (obtained by applying OpenPose~\cite{openpose} on the RGB); (2) the target densepose $D^t$ and body joints $J^t$. We additionally condition on $D^t$ to give the network the flexibility to learn a rough target shape which can not be easily perceived from the joints $J^t$. Thereafter the two extracted bottleneck feature will be concatenated and sent to a series of residual blocks followed by a decoder, yielding the target person mask and segmentation layout $\left(M^t, S^t\right)$. The formulation of this stage can thus be summarized as:
\begin{equation}
    \left(M^t,S^t\right) = f_{csn}\left(R^s \oplus S^s, D^t \oplus J^t\right)
\end{equation}
which is optimized by the combined L1 (for $M^t$) and pixel-level cross-entropy (for $S^t$) loss defined between the predictions and the ground truth from the target frame. The CSN also plays a preparatory role for the following flow estimation stage, where the predicted segmentation can ease the network at capturing spatial structure similarities.

\subsection{Stage 2: Pixel Flow Estimation}\label{subsection:stage2}
As proved in~\cite{han2019clothflow}, the 2D pixel flow mainly reflects structure and texture correspondence between images and thus is object-agnostic, showing its promise of generalization to arbitrary garment types. Motivated by this, we estimate through the PixelFlow Network (PFN) the pixel flow $F^p$ indicating which locations in the target frame the source frame pixels should be mapped to. The inputs of PFN are similar to the CSN, except its target branch additionally receives the predicted segmentations from the CSN.

To give more details, as depicted in Fig.~\ref{fig:framework}.(b), two identical encoders build the contractive steams of PFN, which extract appearance and structure features from the input groups. It might be conceptually simple to directly contrive pixel-wise correspondences from the concatenated downsampling features through chained deconvolutions, as the related work~\cite{han2019clothflow} does. However, in case of large deformation that usually occurs in real world scenes, such a vallina pipeline tends to produce artifacts for the intractable process of the network itself finding the cross-feature association. We hence adopt 
the feature correlation layer of FlowNetCorr proposed in~\cite{Dosovitskiy2015FlowNet} in-between the downsampling and the upsampling parts to impart the network stronger regulation when correlating discrepant features, i.e., a matching score quantizing the feature similarity between images.  
Furthermore, we skip-connect the encoding features to the same-level decoding layers for direct high-level feature transmission that speeds up the learning process. In this way, the decoder now gets a concatenation of the two bottleneck features and their correlation tensor as input, and progressively upsample it to the original image size with a layer-wise flow refinement mechanism, i.e., each flow map estimated by a certain decoding layer will be bilinearly upsampled and concatenated with the flow predicted by the next layer to jointly refine the estimation result. 

The overall process can now be formulated as 
\begin{equation}
   F^p = f_{pfn} \left( R^s \oplus S^s, R^t \oplus S^t \right)  
\end{equation}
With $F^p$, we can map the source frame texture to the target. Note since we train by reposing the same person in different time instance, we thus can self-supervise the flow estimation with cross-frame consistency, i.e., similarity between the mapped texture and the ground truth texture in the target frame. We choose equally weighted perceptual ~\cite{johnson2016perceptual} and L1 losses as the cost function for this stage.

The main difference between our PFN and the related ClothFlow~\cite{han2019clothflow} are three fold: (1) we additionally leverage the rigid densepose $D^t$ as input to partially neutralize the high degrees of flow freedom, which is especially beneficial for tight-dressed persons; (2) we exploit a correlation layer that provides explicit feature matching guidance while in ClothFlow they implicitly explore this via a cascaded flow rectification network, of which the process as aforementioned is relatively hard to control. (3) The ClothFlow warps each encoding feature according to the estimated flow to account for the feature misalignment which we do not, since this has a risk of accumulating errors if the initial predicted flow is inaccurate. With the aid of feature correlation layer and the rich input information, our PFN is preferable to video data taken in the wild. 

\subsection{Stage 3: Garment Transfer with wFlow}\label{subsection:stage3}
The core contribution of this stage is the \textbf{wFlow}. We will first explain the process of obtaining this novel flow, and then dive into detail of the proposed Garment Transfer Network (GFN) as well as the objective functions.

\noindent\textbf{wFlow.} Having the predicted 2D pixel flow from stage 2, we boost its capability by injecting informative 3D SMPL vertex flow. The new blended flow is kinematic-aware that has more pose transfer potential when faced with in-the-wild scenarios, therefore we name it \textit{wFlow} (i.e., flow in the wild). Specifically, we fit SMPL body mesh to $I^s$ and $I^t$ using~\cite{kanazawaHMR18} respectively, and then project the fitted meshes to the densepose representation $\left(D^s, D^t\right)$ in 2D UV space (i.e., image space). Since the SMPL mesh topology is fixed, we can immediately calculate a (2D) vertex flow $F^v$ between $D^s$ and $D^t$, given the barycentric coordinates of each densepose pixel with respect to its unprojected mesh face. We denote the binary mask derived from the vertex flow map as $M^v$, and apply the following formulation to obtain the wFlow $F^w$:
\begin{equation}\label{eqn:blendflow}
    F^w = M^v \odot F^v + (1 - M^v) \odot F^p 
\end{equation}
This formulation has two nice properties. First, the vertex flow component has higher priority to guarantee the correctness of texture mapping for rigid body parts. Second, for non-rigid garment deformation, the pixel flow component will show accordingly its mastery. Thus, by Eq.~\ref{eqn:blendflow}, we factorize the texture mapping into articulation and non-rigid deformation, which we would argue is more flexible than solely using either one component. The ablation study on the wFlow in Sec.~\ref{subsec:ablation} further supports this point.

The $F^w$ warps $I^s$ to $I^s_w$ complying with the target pose, which will then be concatenated with the $S^t$ and the unchanged target person parts $P^t$ (obtained by applying~\cite{liang2018look} on $I^t$), making ready the warped person representation $\left(I^s_w,P^s,S^s\right)$ for the garment transfer network.

\noindent\textbf{Garment Transfer Network (GTN).} As shown in Fig.~\ref{fig:framework}.(c), the GTN has three identical UNet-like generators named $G^B$, $G^S$ and $G^T$, where (1) $G^B$ inpaints the source and target backgrounds; (2) $G^S$ reconstructs the source; and (3) $G^T$ synthesizes the pose transfer result during training. Note at testing, the $G^T$ will run garment transfer. Fundamentally, our GTN follows the overall architecture of ~\cite{Liu_2019_liquid}, but adapts it for high fidelity garment transfer especially loose clothes by virtue of the proposed wFlow.

More specifically, the background inpainting generator $G^B$ takes the dilated-masked source and target background $\left(B^s_m, B^t_m\right)$ as batched input, and outputs the respectively inpainted backgrounds $\left(B^s_o, B^t_o\right)$. Afterwards $B^s_o$ will be added to the inputs of $G^S$ together with the densepose-masked source RGB $I^s_d$ and the source mask $M^s$, all of which are passed through $G^S$ to reconstruct $O^s$ as close to $I^s$. Note, since $D^s$ comes from the garment-less SMPL mesh, some parts of loose garments may be masked out in $I^s_d$. This enforces the $G^s$ to learn texture completion for the exterior part of $M^s$ with respect to $I^s_d$. Concurrently, the $B^t_o$ concatenated with the warped representation $\left(I^s_w+P^t+S^t\right)$ are aggregated by $G^T$ to synthesize the pose transfer result $O^t$ (or, the garment transfer result during testing). In practice, the $G^S$ (resp. $G^T$) first predicts a fusion mask $M^s_f$ (resp. $M^t_f$) and a coarse result $\widetilde{O^s}$ (resp. $\widetilde{O^t}$) and then fuse them with $B^s_o$ (resp. $B^t_o$) to the final result $O^s$ (resp. $O^t$):
\begin{equation}
    \begin{array}{l}
    O^{s(t)} = \widetilde{O^{s(t)}} \odot M^{s(t)}_f + B^{s(t)}_o \odot (1 - M^{s(t)}_f) 
    \end{array}
\end{equation}

It is worth noting that we warp and distillate features of $G^S$ to $G^T$ inspired by~\cite{Liu_2019_liquid}, but has two main differences from it: (1) the features are now transformed by the blended wFlow instead of the single vertex flow as in~\cite{Liu_2019_liquid}, which provides non-rigid deformation ability; (2) in~\cite{Liu_2019_liquid} they focus on transmitting structural and texture information between two generators, while we aim to distillate the \textit{inpainted} features of $G^S$ to $G^T$, which can ease the $G^T$ at inference to inpaint textures of loose garment. This is again the reason why $G^s$ needs the capability of texture completion realized by setting its input to $M^s$ instead of $D^s$ used by the other work. Besides, the $G^T$'s dependency on $G^S$ also shows the necessity of leveraging two generators instead of one ($G^T$-only). Please refer to the supplementary for more detail of this feature distillation operation.

\noindent\textbf{Loss functions.} The training losses of GTN are computed against three products: the fusion mask $M_f^{s(t)}$, the reconstructed $O^s$, and the pose transfer result $O^t$. In particular, we adopt the BCELoss $\mathcal{L}_{BCE}$ for $M^s_f$ and $M^t_f$ under the regularization of Total Variation constraints~\cite{Liu2019tvloss}, which is formulated as (same for $M^s_f$):
$$
  \mathcal{L}_{m}\left(M^t_f\right) =\mathcal{L}_{BCE}\left(M^t_f, M^t\right)+T V\left(M^{t}_f\right),  
$$
where the TV loss in detail is: 
\begin{align*}
    T V(M^{t}_f) &= \sum_{i, j}[(M^{t}_f(i, j)-(M^{t}_f(i-1, j)]^{2} \\
            &+[(M^{t}_f(i, j)-(M^{t}_f(i, j-1)]^{2}.    
\end{align*}
As for the reconstructed $O^s$ and the synthesized $O^t$, we use L1 ($\mathcal{L}_1$) and perceptual loss ($\mathcal{L}_{perc}$)to measure their difference to the ground truth frame images. An adversarial loss based on the Pix2Pix discriminator~\cite{pix2pix2017} is further incorporated for $G^T$, narrowing the distribution gap between the synthesized and the real images. Thus, the total loss of the GTN generators is summarized as:
\begin{equation*}
    \mathcal{L}^G_{gtn} = \mathcal{L}_{m} + \mathcal{L}_{1} + \mathcal{L}_{perc} + \mathcal{L}^G_{adv},
\end{equation*}
where $\mathcal{L}^G_{adv} = \sum{D(O^t,S^t)^2}$ in which the $D$ denoting the discriminator and has its own loss
\begin{equation*}
\mathcal{L}^{D}=\sum\left[D\left(O^t,S^t\right)+1\right]^{2}+\sum\left[D\left(I^t,S^t\right)-1\right]^{2}.
\end{equation*}

\subsection{Cyclic Online Optimization}
\begin{figure}
  \centering
  \includegraphics[width=1.0\hsize]{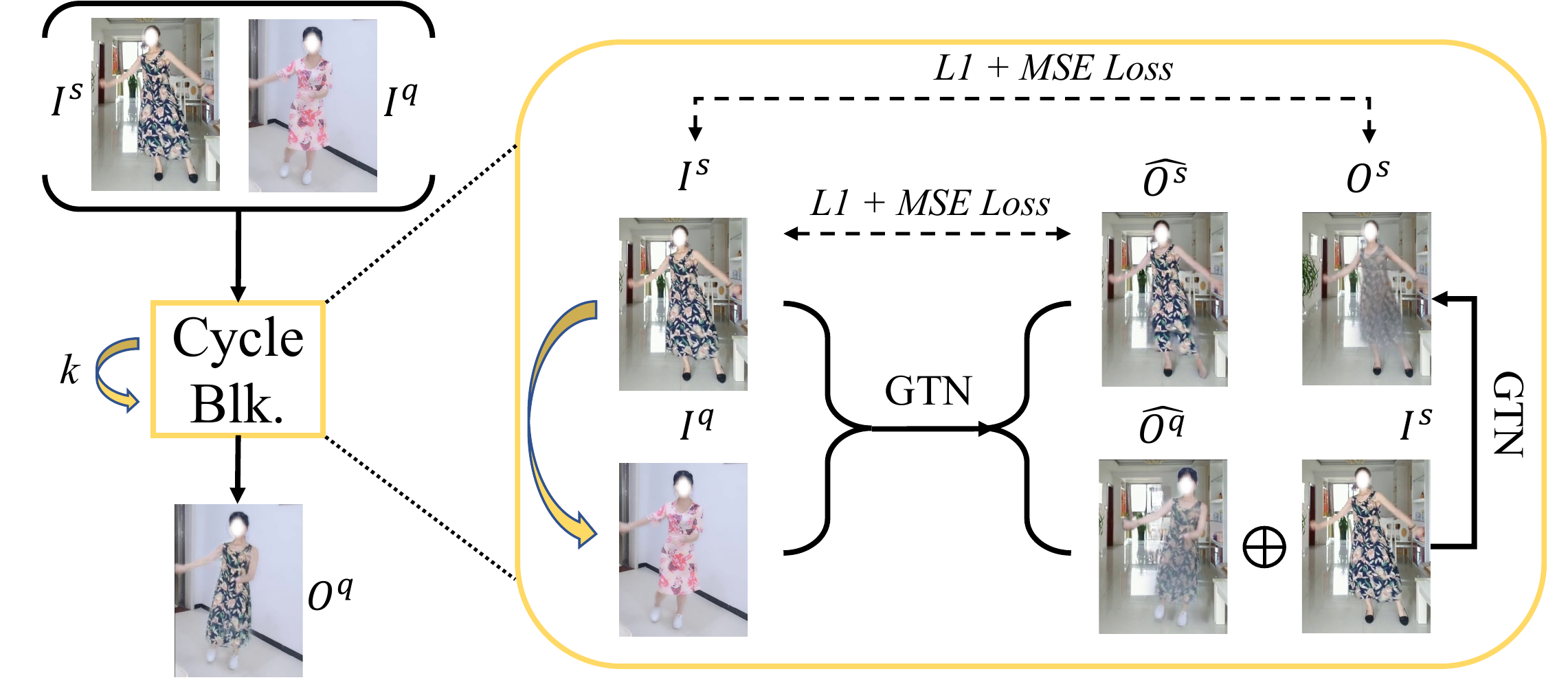}
  \vspace{-6mm}
  \caption{Illustration of the cycle online optimization.} 
  \vspace{-4mm}
  \label{fig:optimization}
\end{figure}

The different setting of training and testing (pose transfer v.s. garment transfer) makes it challenge to directly confront with arbitrary query images especially those low-resolution or indistinct-foreground ones. 
We thus introduce an online optimization depicted in Fig.~\ref{fig:optimization} that works on the cycle consistency to progressively refine the synthesized dressing result whenever the query image quality is unsatisfactory.

In specific, during inference for a pair of source and query images $\left(I^s, I^q\right)$, we pass them $k$ times through the carefully designed \textbf{Cycle Block}, where $k$ is a tunable parameter that trades off the running time and the refinement degree. 
During the first pass, we first transfer the garment of $I^s$ to $I^q$ via the GTN, producing the reconstructed $\widehat{O^s}$ and the intermediate try-on image $\widehat{O^q}$. Thereafter the $\widehat{O^q}$ together with the input $I^s$ will be once more processed by the same GTN, but here the transfer direction reversed ($\widehat{O^q} \rightarrow I^s$), yielding the ``dressed-back'' image $O^s$ where the ``cycle'' closed. The combined L1 and MSE Loss for $\widehat{O^s}$ and $O^s$ with respect to $I^s$ is used to guide this cyclic process. However, at the entrance of second pass, the role of $I^s$ and $I^q$ will exchange: $I^q$ now becomes the ``source'' providing garments while the ``query'' $I^s$ now wants to dress from $I^q$. Intuitively, by exchanging garments repeatedly between a given input pair, we want the GTN to overfit on them at inference. In this way, the output try-on result will be progressively refined to high quality with sharper edges and more realistic textures, largely mitigate the problem of dealing with low-quality input query image.



\section{Experiments}\label{sec:exp}
\begin{figure*}
\begin{center}
\includegraphics[width=1.0\hsize]{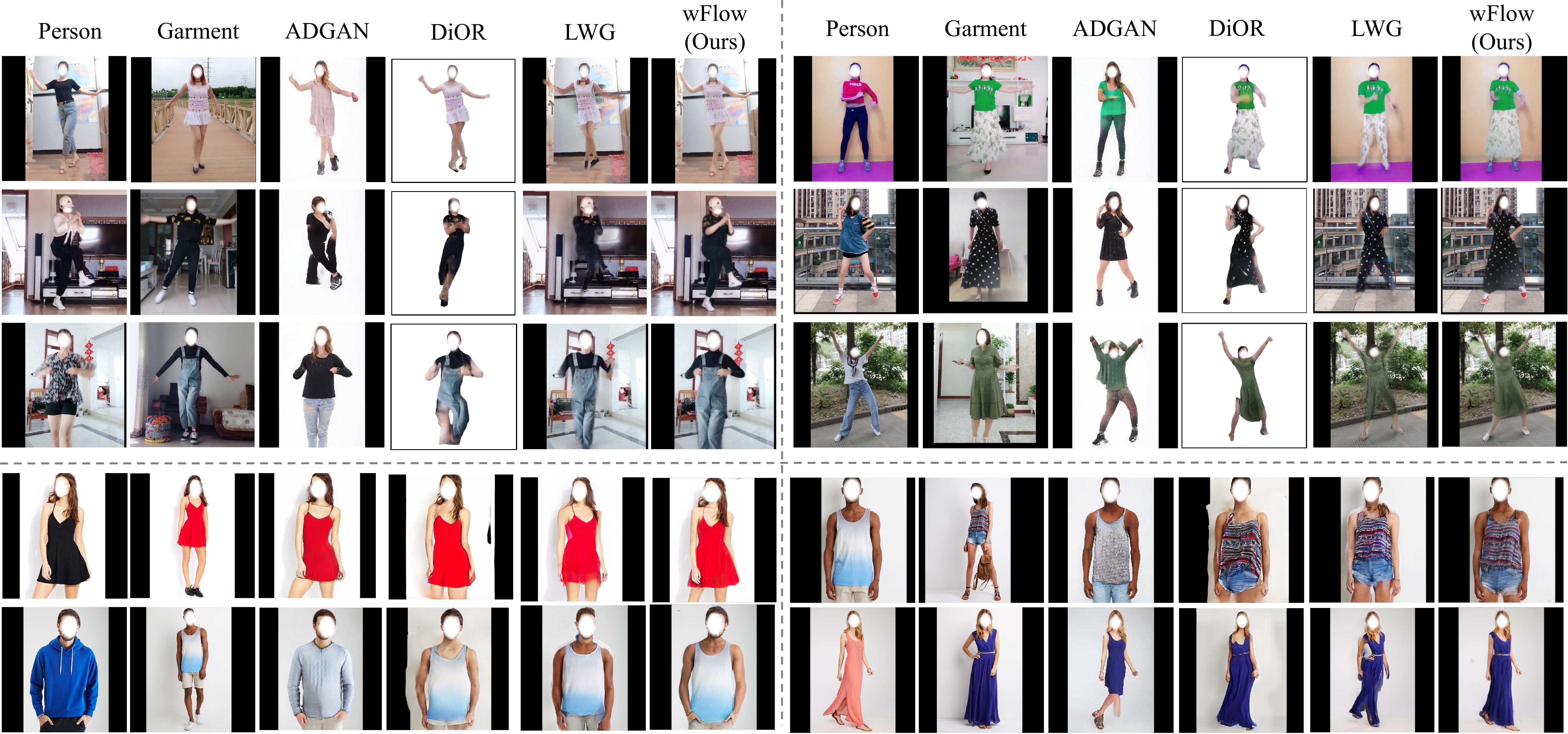}
\end{center}
    \vspace{-4mm}
   \caption{Qualitative comparisons on \textit{Dance50k} (1-3rd rows) and DeepFashion Datset (4-5th rows). The first two columns represent the inputs, while the others are garment transfer results from our method and the other three baselines (LWG~\cite{Liu_2019_liquid}, ADGAN~\cite{men2020adgan} and DiOR~\cite{dior}). Our wFlow contains richer foreground and background texture details and more successfully transfer the loose garments. Please zoom in for more details  and more visual results are provided in supplementary.}
\label{fig:comparison}
\end{figure*}

\begin{figure}
  \centering
  \includegraphics[width=1.0\hsize]{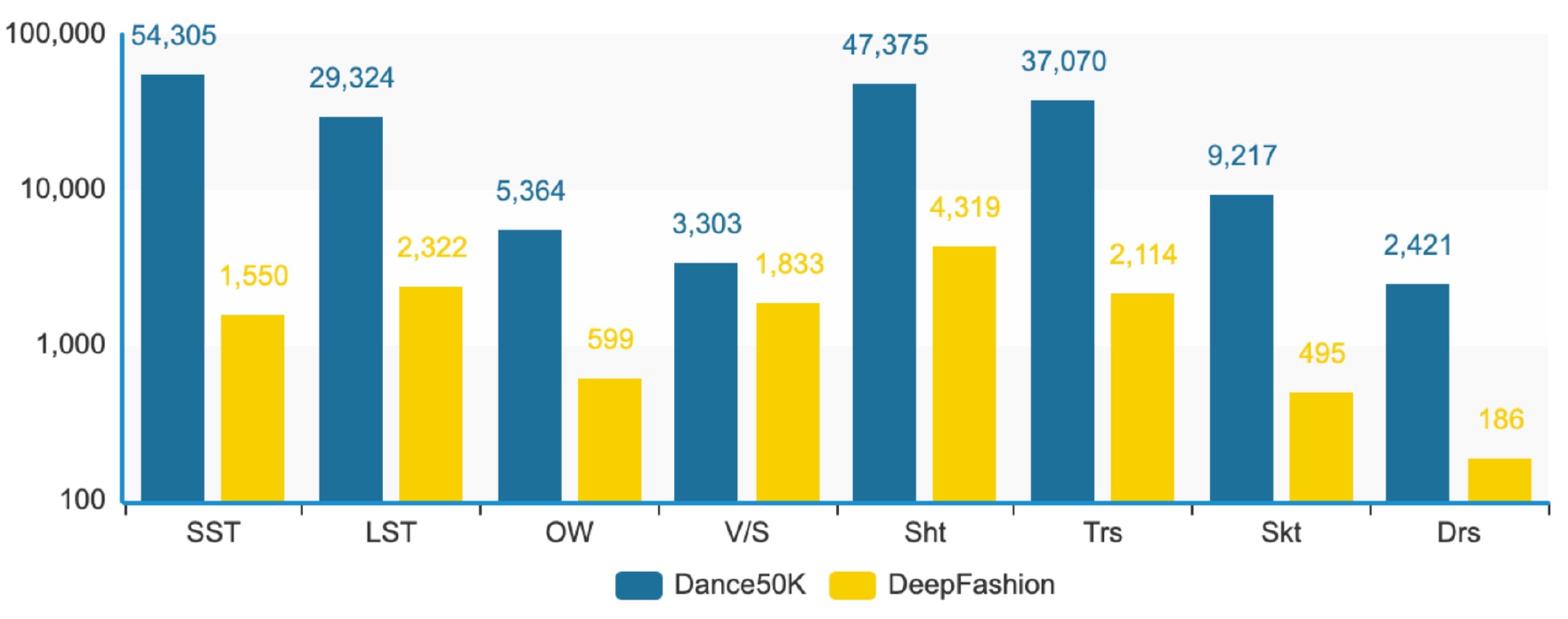}
  \vspace{-4mm}
  \caption{Garment distribution in Dance50k and DeepFashion Dataset. The terms in x-axis respectively denote Short Sleeve Top (SST), Long Sleeve Top (LST), Outwear (OW), Vest/Sling (V/S), Shorts (Sht), Trousers (Trs), Skirts (Skt) and Dress (Drs).} 
  \vspace{-4mm}
  \label{fig:dataset}
\end{figure}
\subsection{Dataset: \textit{Dance50k}}
We learn realistic garment transfer by leveraging a collection of real world dance videos scraped from public internet. The dataset, named \textit{Dance50k}, contains $50,000$ single-person dance sequences (about 15s duration) that feature varying poses and numerous garment types. Fig.~\ref{fig:dataset} plots the garment distribution presented in \textit{Dance50k} and the other popular try-on dataset named DeepFashion~\cite{liuLQWTcvpr16DeepFashion}, revealing ours superiority in garment diversity, which is essential for virtual dressing tasks.
Note, though the \textit{Dance50k} in its modality is very similar to the TikTok Dataset~\cite{Jafarian_2021_CVPR_TikTok}, ours is two orders of magnitude larger than the TikTok (50k v.s. 300), showing the stronger suitability of $Dance50k$ for solving potential human-centric image/video problems that not limit to virtual try-on~\footnote{More detailed description and data examples of \textit{Dance50k} are enclosed in the Supplementary.}.


\subsection{Implementation Details}
We follow a multi-stage training regime for faster convergence and stable training. 
The proposed CSN, PFN and GTN are trained separately for $2.5M$ iterations with a batch size of 4 using the Adam optimizer ($\beta_{1}=0.5$, $\beta_{2}=0.999$), and the learning rate is set to 0.0002. 
We use Pytorch to implement the pipeline and train it on a single NVIDIA V100 GPU~\footnote{We provide the complete model details in the supplementary.}. During training, we uniformly sample 10 frames from each video in \textit{Dance50k} and get a combination of $C_{10}^2$ candidate $(I^s,I^t)$ pairs. To guarantee validation of the flow direction, a simple assertion is further applied on the candidates to ensure that the number of joints in $I^s$ is more than that in $I^t$. Consequently, a total of 949623 image pairs are finally used for training in a generalizable manner of pose transfer. While at testing, the identity of query image $I^q$ can be different with the source $I^s$ and the inference is performed in an end-to-end manner of garment transfer. If the cycle online optimization is further needed for some hard queries, we set the $k$ to $20$ for best time-quality trade-off.  
\subsection{Comparison with Baselines}
\textbf{Evaluation metric.} To fully compare with baselines, we evaluate the performance in two aspects: (1) Fidelity of the pose transfer results; (2) Realism of the synthesized garment transfer images. We adopt the common GAN metrics to measure them respectively, i.e., the Structural SIMilarity index (SSIM)~\cite{Wang2004SSIM} for pose transfer, the Fr$\mathbf{\acute{e}}$chet Inception Distance (FID)~\cite{heusel2017fid} and the Perceptual distance (LPIPS)~\cite{zhang2018perceptual} for garment transfer, as ground truth images are necessary for the SSIM and LPIPS while not for the FID. 
We further evaluate the accuracy of the generated garment shape by computing the Intersection over Union (IoU) between the generated and the real garment silhouette, where larger IoU means the generated shape is more consistent with the real. Note only loose garments (e.g., skirt, dresses) are taken into account for calculating IoU as tight clothing can be modeled as part of the human skin that presents small IoU variance.
We additionally conduct a human evaluation to assess the results (please refer to the supplenmentary for detailed setting of this human evaluation), and besides \textit{Dance50k}, we also report scores on the commonly-used DeepFashion dataset~\cite{liuLQWTcvpr16DeepFashion}.

\begin{table*}[t]
\centering
\begin{tabular}{l c c c c c c c c c c c c c}
  \toprule
  \multicolumn{2}{c}{Dataset}& & \multicolumn{5}{c}{Dance50k} & & \multicolumn{5}{c}{DeepFashion} \\
  \cmidrule{1-2} \cmidrule{4-8} \cmidrule{10-14}
  \multicolumn{2}{c}{Method} & & SSIM $\uparrow$ & FID $\downarrow$ & LPIPS $\downarrow$ & IoU $\uparrow$ & HE $\uparrow$  & & SSIM $\uparrow$ & FID $\downarrow$ & LPIPS $\downarrow$ & IoU $\uparrow$ & HE $\uparrow$ \\
  \cmidrule{1-2} \cmidrule{4-8} \cmidrule{10-14}
  \multicolumn{2}{c}{LWG~\cite{Liu_2019_liquid}} & & 0.891 & 13.080 & 0.107 & 0.484 & 0.271 & & 0.778 & 59.239 & 0.225 & 0.508 & 0.284 \\
  \multicolumn{2}{c}{ADGAN~\cite{men2020adgan}} & & 0.765 & 44.280 & 0.223 & 0.289 & - & & 0.643 & 85.083 & 0.317 & 0.590  & - \\
  \multicolumn{2}{c}{DiOR~\cite{dior}} & & 0.884 & 58.073 & 0.108 & 0.673 & - & & 0.728 & 76.068 & 0.291 & 0.616 & - \\
  \cmidrule{1-2} \cmidrule{4-8} \cmidrule{10-14}
  \multicolumn{2}{c}{\textbf{wFlow (Ours)}} & & \textbf{0.920} & \textbf{8.809} & \textbf{0.090} & \textbf{0.719} & \textbf{0.729} & & \textbf{0.844} & \textbf{57.652} & \textbf{0.187} & \textbf{0.687} & \textbf{0.716} \\
  \bottomrule
\end{tabular}
\caption{Quantitative comparisons to other garment transfer methods, i.e., Liquid Warping GAN~\cite{Liu_2019_liquid}, ADGAN~\cite{men2020adgan} and DiOR~\cite{dior}. Note the SSIMs are reported for pose transfer results while FIDs and LPIPSs are for garment transfer. HE here refers to Human Evaluation and since only LWG and our wFlow can generate background, the HE is reported on these two methods excluding the other two baselines.}
\label{tab:quant}
\end{table*}

\begin{table*}[t!]
\centering
\begin{tabular}{p{0.2cm}p{0.2cm}<{\centering}p{0.2cm}<{\centering}cccccccccc}
\toprule
\multirow{2}{*}[-3pt]{CO} & \multirow{2}{*}[-3pt]{$F^p$} & \multirow{2}{*}[-3pt]{$F^v$} & & \multicolumn{4}{c}{Dance50k} & & \multicolumn{4 }{c}{DeepFashion} \\ 
\cmidrule{5-8} \cmidrule{10-13} 
  &   &   & & SSIM$\uparrow$ & FID$\downarrow$ & LPIPS $\downarrow$ & IoU$\uparrow$ & & SSIM$\uparrow$ & FID$\downarrow$ & LPIPS $\downarrow$ & IoU$\uparrow$ \\
\cmidrule{1-3} \cmidrule{5-8} \cmidrule{10-13} 
\cmark & \cmark & \xmark & & 0.920   & 12.077 &  0.096 & 0.685   & & 0.842   & 63. 136 & 0.195 & 0.648   \\
\cmark & \xmark & \cmark & & \textbf{0.922} & 9.455 & 0.093 & 0.699  & & \textbf{0.847}     & 59.214 & 0.191 & \textbf{0.687}    \\
\xmark & \cmark & \cmark & & 0.920     & 12.106 & 0.099 & 0.709   & & 0.806     & 71.016 & 0.228 &  0.672    \\
\cmidrule{1-3} \cmidrule{5-8} \cmidrule{10-13}
\cmark & \cmark & \cmark & & {0.920}     & \textbf{8.809} & \textbf{0.090} & \textbf{0.719}  & & {0.844}     & \textbf{57.652} & \textbf{0.187} & \textbf{0.687}    \\ \bottomrule
\end{tabular}
\caption{The ablation study on the Pixel Flow ($F^p$), Vertex Flow ($F^v$) and the cycle online optimization (CO).}
\vspace{-3mm}
\label{tab:ablation}
\end{table*}
\noindent\textbf{Qualitative comparison.} As shown in Fig.~\ref{fig:comparison}, we conduct qualitative comparison on $Dance5k$ and DeepFashion~\cite{liuLQWTcvpr16DeepFashion}, with the other three open source state-of-the-art dressing approaches. ADGAN~\cite{men2020adgan} is incapable of fusing the background and fails to correctly preserve the garment attributes. DiOR~\cite{dior} is a versatile model that can handle pose/garment transfer and texture editing in an end-to-end pipeline, but it underperforms for in-the-wild scenes even with the attention-aided flow estimation~\cite{gfla}, producing unrealistic texture distortion. While LWG~\cite{Liu_2019_liquid} genenrates reasonable results, it tends to produce blurred body edges when faced with complex poses and can not model loose clothes. With the powerful wFlow, our garment transfer network can not only handle arbitrary clothing but also seamlessly synthesize the garment transfer foreground.

\noindent\textbf{Quantitative comparison.} As reported in Table~\ref{tab:quant}, our \textbf{wFlow}-based garment transfer network leads all five metrics especially the highlighting FID calculated on the \textit{Dance50k} dataset. The best FID and LPIPS indicates that our wFlow performs better garment transfer on in-the-wild imagery, without sacrificing the performance of high-fidelity pose transfer.
Moreover, the IoU directly shows wFlow can obtain more accurate warping silhouettes, which is crucial for the subsequent texture fusion process.
SSIM measures the structural and brightness similarity between pose transferred images and their ground truth. In our implementation, the vertex flow directly transmits source pixels without changing the pixel value, while the pixel flow result is further fused with an estimated alpha blend mask. As a result, injecting pixel flow into vertex flow will slightly change the brightness of transformed pixels, which will lower a bit the SSIM score for skin-tight clothes favored by the vertex flow. This is why SSIM drops with full method configuration. However, as we attend to in-the-wild garment transfer, combining these two types of flow can significantly improve the overall visual performance especially for loose outfits. This is also supported by the visualized results and other quantitative metrics including the human evaluation, as most volunteers appreciate the superiority of our wFlow in recovering sharp texture and preserving intact garment shape.

\subsection{Ablation Study}\label{subsec:ablation}
\begin{figure}
  \centering
  \includegraphics[width=1.0\hsize]{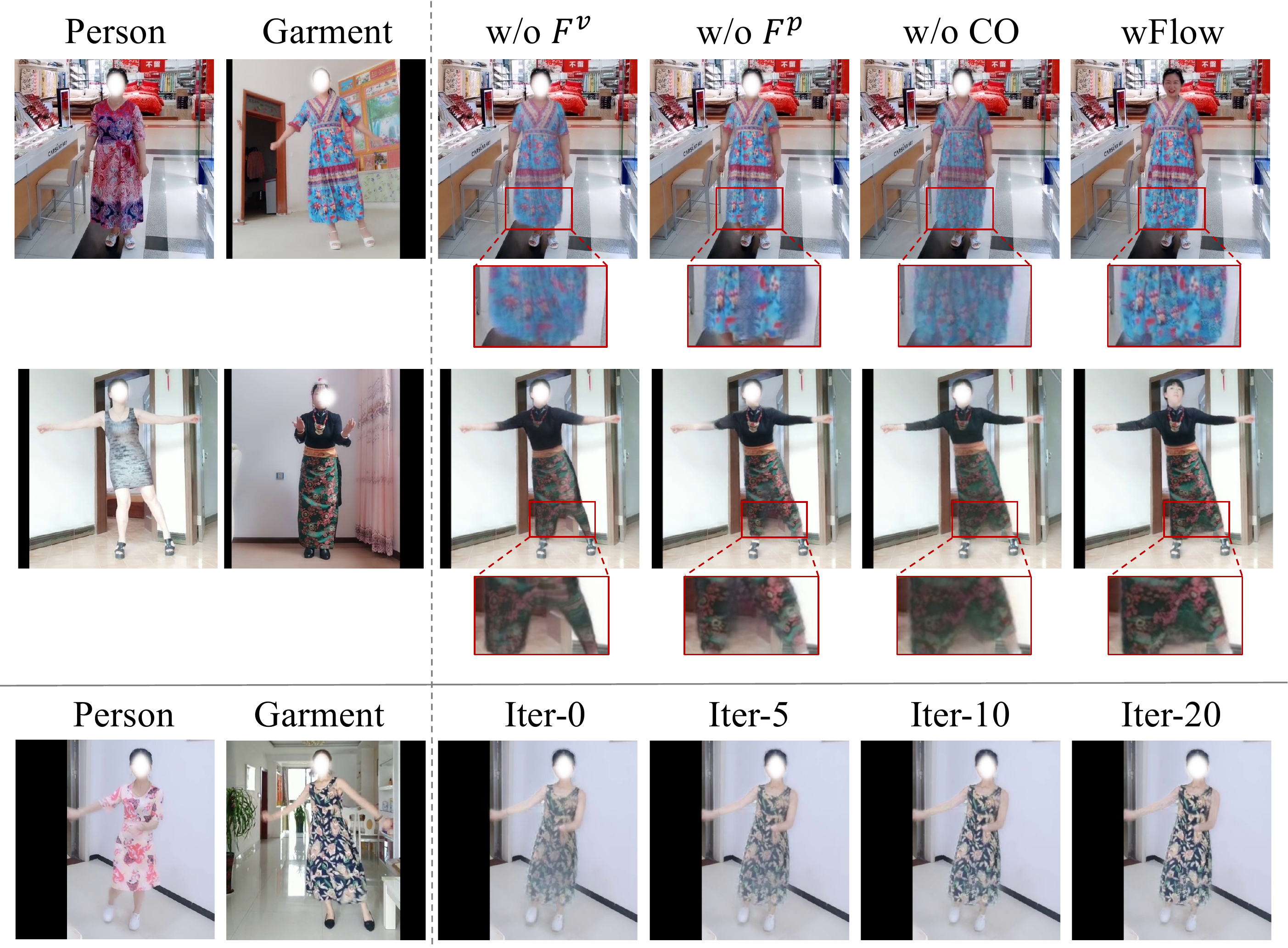}
  \vspace{-6mm}
  \caption{Ablation studies on the wFlow (upper part) and visualized effect of the online cycle optimization (lower part). Please zoom in for more details.} 
  \vspace{-5mm}
  \label{fig:ablation}
\end{figure}
We conduct three ablation studies to analyze the impact of the wFlow and the online Cycle Optimization. The corresponding evaluations are summarized in Table~\ref{tab:ablation}. Solely using $F^p$ (resp. $F^v$) leads to inferior capability of capturing global texture features (resp. human kinematics information), harming either the SSIM or the FID score (Table~\ref{tab:ablation}, 1st-2nd row). since the cyclic online optimization helps refining overall quality of the synthesized images, removing it will also affect negatively (Table~\ref{tab:ablation}, 3rd row).
By leveraging all the three building blocks, the full model (Table~\ref{tab:ablation}, 4th row) outperforms all metrics except SSIM reported for pose transfer. With wFlow, the key FID scores (reported for garment transfer) are improved by a large margin on both datasets, showing the success of our purposeful architecture design that pursues the in-the-wild dressing achievement. 


Accompanying the Table~\ref{tab:ablation}, the upper part of fig.~\ref{fig:ablation} illustrates consequences of ablating the wFlow. Concretely, sorely using $F^p$ or lacking the online CO procedure usually leads to blur garment texture, while sorely using $F^v$ can not guarantee the consistency of garment texture and accuracy of garment shape. Furthermore, the lower part of fig.~\ref{fig:ablation} demonstrates that the fidelity of garment texture can be progressively enhanced during the online CO procedure.



\section{Conclusion}
In this work, we propose a novel multi-stage garment transfer network that performs robustly on in-the-wild imagery. By leveraging easily accessible dance videos, our model predicts a blended flow called wFlow integrating both 2D and 3D body information to map garment textures between different persons. To further enhance the synthesized quality of substandard queries, a novel cyclic optimization is incorporated to iteratively refine the dressing result. We also envision the new dataset, \textit{Dance50k}, can be used to facilitate related human-centric research areas that not limit to virtual try-on. The architecture presented in this paper provides a practical and reliable solution for real world virtual dressing application.

\section{Acknowledgement}
This work was supported in part by National Natural Science Foundation of China (NSFC) under Grant No.61976233, Guangdong Province Basic and Applied Basic Research (Regional Joint Fund-Key) Grant No.2019B1515120039, Guangdong Outstanding Youth Fund (Grant No.2021B1515020061), Shenzhen Fundamental Research Program (Project No.RCYX20200714114642083, No.JCYJ20190807154211365).

{\small
\bibliographystyle{ieee_fullname}
\bibliography{egbib}
}

\end{document}